\documentclass[runningheads]{llncs}
\usepackage[T1]{fontenc}
\usepackage{graphicx}
\usepackage{hyperref}
\usepackage{color}

\newcommand*\samethanks[1][\value{footnote}]{\footnotemark[#1]}
\usepackage[table]{xcolor}

\usepackage{float}

\usepackage{etoolbox}
\apptocmd{\sloppy}{\hbadness 10000\relax}{}{}

\usepackage{tabularray}

\usepackage{graphicx}

\usepackage{hyperref}
\usepackage[T1]{fontenc}

\usepackage{numprint}
\usepackage{newclude}

\usepackage{cleveref}

\begin{document}
\title{Deep Quality Estimation: Creating Surrogate Models for Human Quality Ratings}
\titlerunning{Deep Quality Estimation}

% without orcid
\author{
Florian Kofler\inst{1,2,3,7} \and
Ivan Ezhov\inst{1,2} \and
Lucas Fidon\inst{4} \and
Izabela Horvath\inst{1,5} \and
Ezequiel de la Rosa\inst{8,1} \and
John LaMaster\inst{1,12} \and
Hongwei Li\inst{1,12} \and
Tom Finck\inst{5,6} \and
Suprosanna Shit\inst{1,2} \and
Johannes Paetzold\inst{1,2,5,6} \and
Spyridon Bakas\inst{9,10,11} \and
Marie Piraud\inst{7} \and
Jan Kirschke\inst{3} \and
Tom Vercauteren\inst{4} \and
Claus Zimmer\inst{3} \and
Benedikt Wiestler\inst{3} \thanks{contributed equally as senior authors} \and
Bjoern Menze\inst{1,12} \samethanks
}

\authorrunning{F. Kofler et al.}

\institute{
% 1
Department of Informatics, Technical University Munich, Germany \and
% 2
TranslaTUM - Central Institute for Translational Cancer Research, Technical University of Munich, Germany \and
% 3
Department of Diagnostic and Interventional Neuroradiology, School of Medicine, Klinikum rechts der Isar, Technical University of Munich, Germany \and
% 4
School of Biomedical Engineering \& Imaging Sciences, King's College London, United Kingdom \and
% 5
Insitute for Tissue Engineering and Regenerative Medicine, Helmholtz Institute Munich (iTERM), Germany \and
% 6
Imperial College London, South Kensington, London, United Kingdom \and
% 7
Helmholtz AI, Helmholtz Zentrum München, Germany \and
% 8
icometrix, Leuven, Belgium \and
% 9
Center for Biomedical Image Computing and Analytics (CBICA), University of Pennsylvania, Philadelphia, PA, USA\and
% 10
Department of Radiology, Perelman School of Medicine, University of Pennsylvania, Philadelphia, PA, USA\and
% 11
Department of Pathology and Laboratory Medicine, Perelman School of Medicine, University of Pennsylvania, Philadelphia, PA, USA\and
% 12
Department of Quantitative Biomedicine, University of Zurich, Switzerland
}

% Florian Kofler ( TUM ) < florian.kofler@tum.de> 
% Ivan Ezhov ( TUM ) < ivan.ezhov@tum.de> 
% Lucas Fidon ( King's College London ) < lucas.fidon@kcl.ac.uk> 
% Izabela Horvath ( TUM ) < izabela.horvath@tum.de> 
% Ezequiel de la Rosa ( Technical University of Munich ) < ezequiel.delarosa@icometrix.com> 
% John LaMaster ( Technical University of Munich ) < jlamaste@gmail.com> 
% Hongwei Li ( Technical University of Munich ) < hongwei.li@tum.de> 
% Tom Finck ( Technische Universität München ) < tom.finck@tum.de> 
% Suprosanna Shit ( TUM ) < suprosanna.shit@tum.de> 
% Johannes Paetzold ( TUM ) < johannes.paetzold@tum.de> 
% Spyridon Bakas ( University of Pennsylvania ) < sbakas@upenn.edu> 
% Marie Piraud ( Helmholtz Zentrum Muenchen ) < marie.piraud@helmholtz-muenchen.de> 
% Jan Kirschke ( Abteilung für diagnostische und interventionelle Neuroradiologie, Klinikum rechts der Isar ) < jan.kirschke@tum.de> 
% Tom Vercauteren ( King's College London ) < tom.vercauteren@kcl.ac.uk> 
% Claus Zimmer ( TUM Munich ) < claus.zimmer@tum.de> 
% Benedikt Wiestler ( TUM ) < b.wiestler@tum.de> 
% Bjoern Menze ( TUM ) < bjoern.menze@tum.de>

% author emails (no particular order)
% florian.kofler@tum.de
% ivan.ezhov@tum.de
% lucas.fidon@kcl.ac.uk
% ezequiel.delarosa@icometrix.com

% johannes.paetzold@tum.de
% hongwei.li@tum.de
% izabela.horvath@tum.de
% suprosanna.shit@tum.de
% jlamaste@gmail.com
% tom.finck@tum.de

% seniors
% sbakas@upenn.edu
% marie.piraud@helmholtz-muenchen.de 
% jan.kirschke@tum.de
% tom.vercauteren@kcl.ac.uk 
% claus.zimmer@tum.de
% b.wiestler@tum.de
% bjoern.menze@tum.de

\maketitle              % typeset the header of the contribution

\begin{abstract}
Human ratings are abstract representations of segmentation quality.
To approximate human quality ratings on scarce expert data, we train surrogate quality estimation models.
We evaluate on a complex multi-class segmentation problem, specifically glioma segmentation, following the BraTS annotation protocol.
The training data features quality ratings from 15 expert neuroradiologists on a scale ranging from 1 to 6 stars for various computer-generated and manual 3D annotations.
Even though the networks operate on 2D images and with scarce training data, we can approximate segmentation quality within a margin of error comparable to human intra-rater reliability.
Segmentation quality prediction has broad applications.
While an understanding of segmentation quality is imperative for successful clinical translation of automatic segmentation quality algorithms, it can play an essential role in training new segmentation models.
Due to the split-second inference times, it can be directly applied within a loss function or as a fully-automatic dataset curation mechanism in a federated learning setting.
\keywords{
automatic quality control \and
quality estimation \and
segmentation quality metrics \and
glioma \and
BraTS
}

\end{abstract}

\section{Introduction}
Large and long-standing community challenges,  such as BraTS \cite{bakas2019identifying}, have created a multitude of fully-automatic segmentation algorithms over the years.
To fully exploit the potential of these task-specific algorithms, be it for clinical or scientific purposes, it is essential to understand the quality of their predictions and account for segmentation failures.
\footnote{
Models often output uncertainty levels to account for this.
However, we believe the judgment of external entities is more trustworthy for quality assurance purposes.
Remarkably, while this separation of concerns is a well-established practice in other fields, it remains largely ignored in machine learning.
For instance, imagine a world where aircraft pilots could self-certify their ability to fly.
}

As individual segmentation quality metrics are only able to cover isolated aspects of segmentation quality, most segmentation challenges evaluate on a combination of metrics \cite{maier2018rankings}.
Human expert quality ratings have become a prominent tool to complement conventional analysis, among others \cite{li2019diamondgan,moller2020reliable,thomas2022improving} for segmentation outputs \cite{kofler2021using}.
In contrast, to narrowly defined quality metrics, these more holistic measures capture various quality aspects.
However, they are prohibitively expensive regarding data acquisition times and financials to be deployed regularly.

\noindent\textbf{Related work:}
Therefore, previous research tried to approximate segmentation performance by other means.
Reverse classification accuracy (RCA) has been proposed to evaluate segmentation accuracy for applications of multi-organ segmentation in magnetic resonance imaging (MRI) \cite{valindria2017reverse} and cardiac MRI segmentation~\cite{robinson2019automated}.
The authors used the maximum predicted segmentation quality metric estimated by a multi-atlas-based registration method as a proxy to estimate the true quality metric.
However, the application of RCA is so far limited to organ segmentation.
Further, the analysis is restricted to established segmentation quality metrics such as DSC and Hausdorff distance.
It is unclear how it would generalize to expert scoring and to lesion segmentation.

A method to estimate the Dice score of an ensemble of convolutional neural networks (CNNs) for segmentation has also been proposed~\cite{hann2021deep}.
They propose to train a linear classifier regression model to predict the Dice score for every CNN in the ensemble.
Here, the Dice scores for every pair of segmentation predictions of the models in the ensemble serve as input for the regressor.
Further, Audelan et al. proposed an unsupervised learning method for predicting DSC using Bayesian learning \cite{audelan2019unsupervised}.

The above methods are coupled with segmentation algorithms for the estimation of segmentation quality metrics.
Unlike this, Fournel et al. predict the segmentation quality metric directly from the input image and segmentation map using a CNN \cite{fournel2021medical}.
Similarly, it is possible to predict an ensemble's segmentation performance from discord between individual segmentation maps, even when ignoring the image data \cite{kofler2021robust}.

The BraTS challenge \cite{menze2014multimodal} features the multi-class segmentation problem of glioma segmentation.
Distinguishing between \emph{enhancing tumor}, \emph{necrosis}, and \emph{edema} is a complex subtask scattered over multiple imaging modalities.
It is evaluated using Sørensen–Dice coefficient (DSC) and Hausdorff distance (HD) for the \emph{whole tumor}, \emph{tumor core} and \emph{enhancing tumor} channels.
Previous research  revealed that BraTS tumor segmentation algorithms typically perform well when monitoring established segmentation quality ratings such as DSC, HD, and others.
However, when they fail, they fail spectacularly \cite{fidon2021generalized,bakas2019identifying, kofler2020brats, kofler2021robust,menze2014multimodal}.
These findings reflecting multiple established segmentation quality metrics are supported by surveys with expert neuroradiologists \cite{kofler2021using}.
Here, the multi-faceted concept of segmentation quality was condensed to a single expert quality rating.
To this end, BraTS glioma segmentation is a good candidate for studying how a holistic rating can complement or replace them. 

\noindent\textbf{Contribution:}
In contrast to the above-mentioned methods, which predict narrowly defined quality metrics such as DSC or HD, we focus on the approximation of more holistic expert neuroradiologists' ratings.
We build surrogate regression models for these abstract human quality ratings to estimate the segmentation quality of MICCAI BraTS segmentation algorithms.
A sophisticated augmentation pipeline compensates for the scarce 2D training data available.
Despite these obstacles, our model manages to create robust estimates for 3D segmentation quality on an internal and external test set.
While our models are agnostic to the segmentation method and are even compatible with 2D manual segmentations, split-second inference times enable broad downstream applications in scientific and clinical practice.

\section{Methods: Network Training}
\label{sec:methods}
We train multiple regression networks to approximate the human quality ratings.

\noindent\textbf{Segmentation quality rating:}
We use human quality ratings provided by Kofler et al. \cite{kofler2021using}.
In this study, expert radiologists rated the quality of glioma segmentations in two experiments.
In the first experiment, 15 neuroradiologists rated the segmentations' center of mass for 25 exams with four different segmentations from axial, sagittal, and coronal views, resulting in 300 trials.
The experiment featured one manual and three computer-generated segmentations.
In the second experiment, three neuroradiologists rated another 50 exams with one manual and five computer-generated segmentations only on axial views, again resulting in 300 trials.
The rating scale ranges from 1 star for very bad to 6 stars for very good segmentations.

\begin{figure}[h!]
\includegraphics[width=\textwidth]{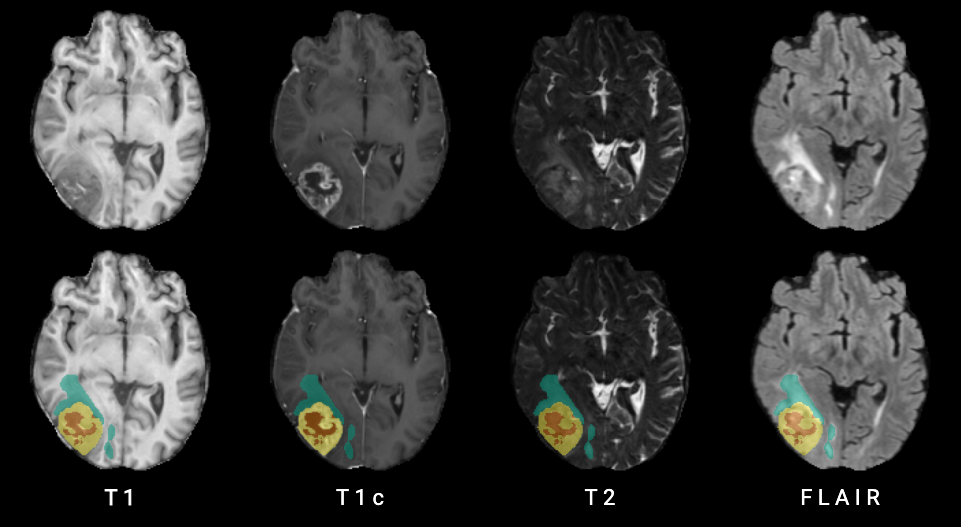}
\caption{
Example inputs for the CNN training - axial center of mass slices of a glioma exam.
Besides the illustrated T1, T1c, T2, and FLAIR MR sequences, the tumor segmentations are supplied to the network.
The different tumor tissue types are visualized in colors:
Red: \emph{necrosis}; Yellow: \emph{enhancing tumor}; Green: \emph{edema}.
For BraTS label encoding, the \emph{enhancing tumor} is encoded in a binary label channel, while a second \emph{tumor core} channel is formed by combining \emph{enhancing tumor} and \emph{necrosis} and a third \emph{whole tumor} channel is calculated by combining all three tumor tissue labels.
}
\label{fig:inputs}
\end{figure}

\noindent\textbf{Network input and output:}
To predict the above-mentioned quality rating, the network receives the four MR modalities, namely T1, T1c, T2, and FLAIR.
The tumor segmentations are either supplied in a single label channel encoding the different tumor tissues or with three label channels following BraTS annotation concepts, as illustrated by \Cref{fig:inputs}.
We try this style of encoding the tumor segmentations as this approach has been proven successful for training BraTS segmentation algorithms \cite{bakas2019identifying}.

\noindent\textbf{Training constants:}
The dataset is randomly split into 80 percent (60) of the 75 exams for training and 20 percent (15) for testing.
Batch size is kept constant at \emph{80} and learning rate at \emph{1e-3}.
To compensate for the scarce training data, we employ a heavy augmentation pipeline featuring Gaussian noise, flips, and random affine plus elastic transformations.
Additionally, we use \emph{batchgenerators} \cite{isensee_fabian_2020_3632567} to augment with contrast, brightness, gamma, low resolution, and rician noise.
Further, we simulate MR artifacts with \emph{TorchIO} \cite{2021torchio}, specifically motion, ghosting, spikes, and bias fields.
We employ a Mean Square Error (MSE) loss for all training runs to especially penalize far-off predictions.
The human quality ratings serve as reference annotations for the loss computations.
Due to the scarcity of training data, we do not conduct model selection and use the last checkpoint after \emph{500} epochs of training across all training runs.

\noindent\textbf{Training variations:}
Between training runs we experiment with three optimizers, namely Ranger21 \cite{wright2021ranger21}, \href{https://pytorch.org/docs/stable/generated/torch.optim.AdamW.html}{AdamW}, and \href{https://pytorch.org/docs/stable/generated/torch.optim.SGD.html}{SGD} with a momentum of $0.95$.
We use \href{https://docs.monai.io/en/stable/networks.html\#densenet121}{DenseNet121} and \href{https://docs.monai.io/en/stable/networks.html\#densenet201}{DenseNet201} \cite{huang2017densely} to investigate whether the performance profits from more trainable parameters.
Moreover, we try a channel-wise min/max, and a \emph{nn-Unet} \cite{isensee_fabian_2020_3632567} inspired percentile-based normalization using the 0.5 and 99.5 percentiles for minimum and maximum, respectively.

\noindent\textbf{Software:}
All computations happen with \emph{NVIDIA Driver v470.103.01}, \emph{CUDA v11.4} with \emph{PyTorch 1.9.0}.
The networks are implemented via \emph{MONAI} \cite{the_monai_consortium_2020_4323059} \emph{0.6.0}.
Segmentation metrics are computed with \emph{pymia 0.3.1} \cite{jungo2021pymia} and regression metrics via \emph{scikit-learn 0.24.2}.

\noindent\textbf{Hardware:}
All computations take place on a small workstation with an \emph{8-core Intel(R) Xeon(R) W-2123 CPU @ 3.60GHz} with \emph{256GB RAM} and a \emph{NVIDIA Quadro P5000} GPU.

\noindent\textbf{Computation times:}
A training run takes approximately two hours.
The above machine infers four 3D exams per second, including mass computation.
Without the extraction of 2D slices, the inference performance increases to 25 per second.
Note that the implementation is not fully-optimized for computation time, as the split-second inference times are in no way impeding our purposes.

\noindent\textbf{Memory consumption:}
With the \emph{batch size} of 80 we use most of the 16gb CUDA memory of the \emph{NVIDIA Quadro P5000} GPU.

\section{Evaluation Experiments}
We conduct two experiments to evaluate the performance of our models.
In the first experiment we identify and evaluate our best performing model.
In the second experiment we validate its generalization capabilities on an external test set.
In both experiments, we generate network predictions for the axial, coronal, and sagittal views and compute a mean rating for each exam.

\subsection{Internal evaluation experiment}
To investigate which of the six hyperparameter combinations works best by evaluating on an internal test set.

\noindent\textbf{Dataset:}
We use the previously held-back 20 percent of the training data for evaluation, as described in \Cref{sec:methods}.

\noindent\textbf{Procedure:}
We run inference on the test set for each model.
Following, we evaluate using established regression metrics, namely mean absolute error (MAE), root mean square error (RMSE).
We further compute Pearson r (r) to measure the linear correlation between network predictions and ground truth labels.

\noindent\textbf{Results:} \Cref{tab:results} illustrates the performance differences between training runs.
We observe that the simple DenseNet121 trained with percentile-based normalization, Ranger21 optimizer, and BraTS label encoding approximates the human rating best.

\begin{table}[htbp]
\caption{
    Training results for different training parameters.
    We report mean absolute error (MAE), root mean square error (RMSE), and Pearson r (r).
    The selected model is highlighted in pink.
    }
\nprounddigits{2}
\npdecimalsign{.}
\centering
% \scriptsize
\begin{tblr}{colspec={lcr},colspec={|Q[2,c] | Q[1.5,c] Q[1.5,c] Q[1,c] | Q[0.5,c] Q[0.5,c] Q[0.5,c]|}}
% \begin{tabular}{rrlrrrr}
  \hline
\textbf{architecture} & \textbf{optimizer} & \textbf{normalization} & \textbf{labels} & \textbf{MAE} & \textbf{RMSE} & \textbf{r}\\ 
  \hline
    \href{https://docs.monai.io/en/stable/networks.html\#densenet121}{DenseNet121} & \href{https://pytorch.org/docs/stable/generated/torch.optim.SGD.html}{SGD} & \href{https://docs.monai.io/en/latest/transforms.html\#scaleintensityrangepercentiles}{percentile} &   tissue &   $\numprint{0.6456292410691579}$ &    $\numprint{0.8787613280718022}$ & $\numprint{0.6448052760689299}$  \\
    \href{https://docs.monai.io/en/stable/networks.html\#densenet121}{DenseNet121} & \href{https://pytorch.org/docs/stable/generated/torch.optim.AdamW.html}{AdamW} & \href{https://docs.monai.io/en/latest/transforms.html\#scaleintensityrangepercentiles}{percentile} &   tissue &   $\numprint{0.5943944454193115}$ &    $\mathbf{\numprint{0.7998518862868392}}$ & $\numprint{0.6176082350140887}$  \\
    \href{https://docs.monai.io/en/stable/networks.html\#densenet121}{DenseNet121} & \href{https://github.com/lessw2020/Ranger21}{Ranger21} & \href{https://docs.monai.io/en/latest/transforms.html\#scaleintensityrangepercentiles}{percentile} &   tissue &   $\numprint{0.5988505641619365}$ &    $\numprint{0.8304728798097425}$ & $\numprint{0.5820392505202615}$  \\
    \href{https://docs.monai.io/en/stable/networks.html\#densenet201}{DenseNet201} & \href{https://github.com/lessw2020/Ranger21}{Ranger21} & \href{https://docs.monai.io/en/latest/transforms.html\#scaleintensityrangepercentiles}{percentile} &   tissue &   $\numprint{0.5912021636962891}$ &    $\mathbf{\numprint{0.7955717833939066}}$ & $\numprint{0.6258785246508369}$  \\
    \href{https://docs.monai.io/en/stable/networks.html\#densenet121}{DenseNet121} & \href{https://github.com/lessw2020/Ranger21}{Ranger21} & \href{https://docs.monai.io/en/latest/transforms.html\#normalizeintensity}{min.max} &   BraTS &   $\numprint{0.5721432904402415}$ &    $\numprint{0.8219236174495188}$ & $\numprint{0.6096361711959628}$  \\

    \hline
    \SetRow{magenta7}
    \href{https://docs.monai.io/en/stable/networks.html\#densenet121}{DenseNet121} & \href{https://github.com/lessw2020/Ranger21}{Ranger21} & \href{https://docs.monai.io/en/latest/transforms.html\#scaleintensityrangepercentiles}{percentile} &   BraTS &   $\mathbf{\numprint{0.5111889163653056}}$ &    $\mathbf{\numprint{0.7964592549345952}}$ & $\mathbf{\numprint{0.6631433019083274}}$  \\
    \hline
\end{tblr}
    \label{tab:results}
\end{table}

\begin{figure}[H]
\includegraphics[width=\textwidth]{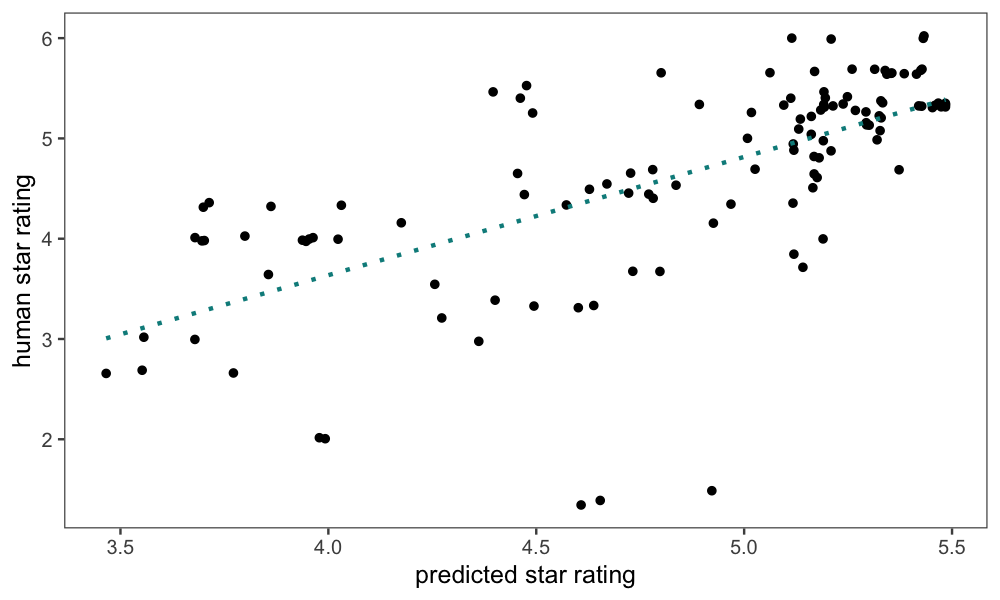}
\caption{
Scatter plot: Network predicted vs. average human star rating.
Even though we have only scarce training data, the model can approximate the human segmentation quality rating quite well.
The dotted cyan line symbolizes a linear model fitted through the data.
We observe a Pearson r of 0.66.
}
\label{fig:network}
\end{figure}

\Cref{fig:network} visualizes the model's predictions compared to the averaged human star ratings.

\begin{figure}[h]
\includegraphics[width=0.95\textwidth]{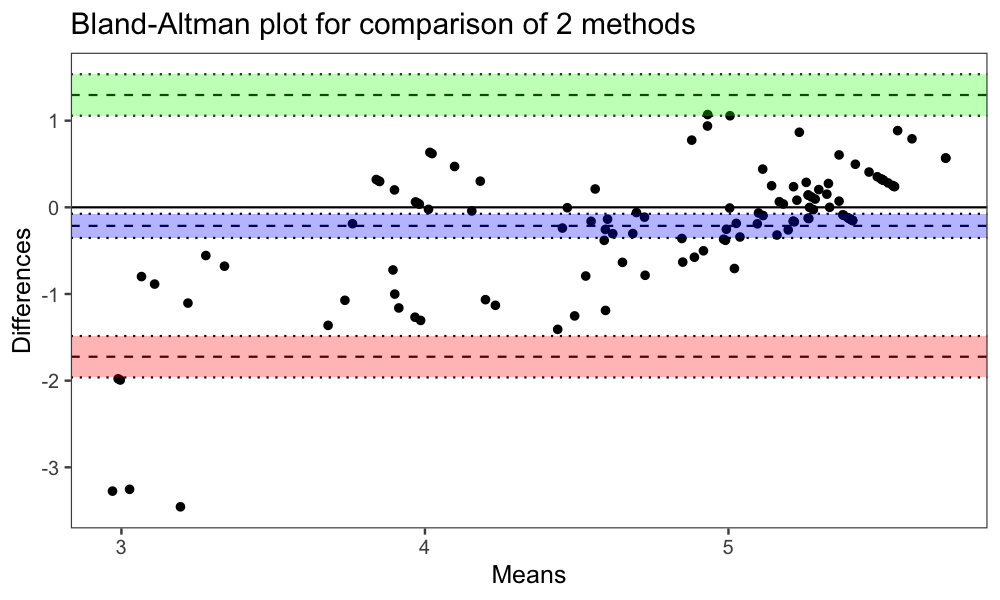}
\caption{
Bland-Altman plot: Network predictions vs. human star rating.
The model reveals higher prediction accuracy for better-quality segmentations.
This is not surprising given that human raters also display higher agreement for such cases and that these are better represented in the training data, compare \Cref{fig:pointo}.
}
\label{fig:blanda}
\end{figure}

\begin{figure}[h!]
\includegraphics[width=1.0\textwidth]{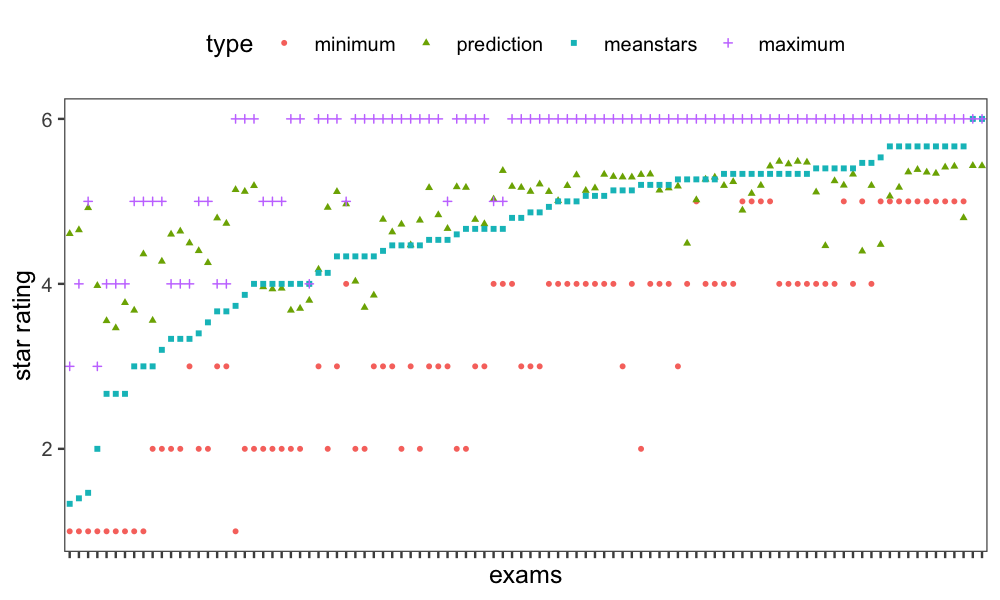}
\caption{
Network predictions vs. minimum, mean, and maximum human rating.
For lower quality segmentations, human raters disagree more.
In most cases, the network's predictions are in the range or close to the human ratings.
As already visible in \Cref{fig:blanda}, the network tends to overestimate the quality of bad segmentations. However, they are still assigned systematically lower scores.
In practice, this difference is sufficient to distinguish between good and bad segmentations by employing a simple thresholding operation.
}
\label{fig:pointo}
\end{figure}

According to the scatter plot, illustrated in \Cref{fig:network}, the model performs more accurately for better segmentations.
This is also reflected by a Bland-Altman plot, see \Cref{fig:blanda}.
It is important to note that the variance in human quality assessment also increases for lower quality segmentations, see \Cref{fig:pointo}.

\subsection{External evaluation experiment}
To better understand our model's generalizability, we further evaluate an external dataset.

\noindent\textbf{Data set:}
The dataset features manual annotations for 68 exams generated by two expert radiologists in consensus voting.
It includes 15 high-grade glioma (GBM) and 13 low-grade glioma (LGG) from \emph{University Hospital rechts der Isar}.
Furthermore, 25 GBM and 15 LGG from the publicly available Rembrandt dataset \cite{gusev2018rembrandt} are added to the analysis.
We obtain five segmentations from BraTS algorithms and four fusions from BraTS Toolkit \cite{kofler2020brats}.
This way, we have a total of 612 segmentations to evaluate.

\noindent\textbf{Procedure:}
We select the best model obtained from the first experiment.
We feed 2D views of the 3D augmentations' center of mass to the network to obtain an axial, sagittal, and coronal quality rating.
As there are no human quality ratings for this dataset, we measure segmentation performance using established quality metrics, namely Sørensen–Dice coefficient (DSC) and surface Dice coefficient (SDSC).

\begin{figure}[H]
\includegraphics[width=1.0\textwidth]{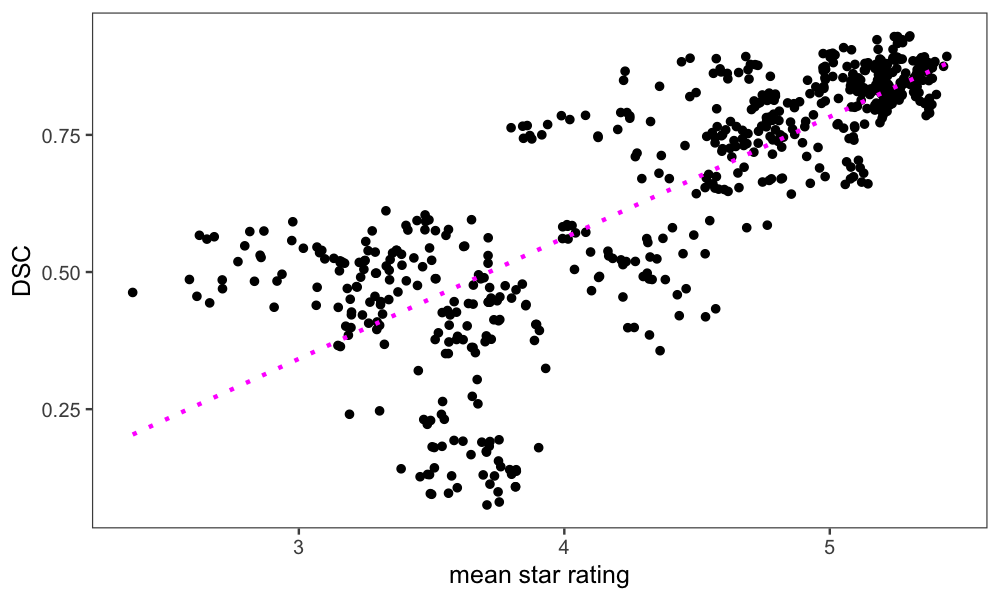}
\caption{
Scatter plot: Predicted star rating averaged across views vs. DSC for 612 segmentations.
Even though we observe moderate heteroscedasticity, a linear model is able to describe the data well, as illustrated by the dotted pink line.
We observe a Pearson r of 0.79 between the variables, meaning our model can predict the segmentation quality as measured by DSC quite well.
Again we observe more accurate predictions for better segmentations.
}
\label{fig:scatter}
\end{figure}

\noindent\textbf{Results:}
We find a strong correlation between the quality ratings predicted by the network and DSC, see \Cref{fig:scatter}.
We observe a Pearson r of 0.75 for the axial, 0.76 for the coronal, and 0.77 for the sagittal view, while the averaged rating across views has a 0.79 correlation.
This is supported by a high correlation between the mean rating and SDSC (Pearson r: 0.85), suggesting that the model generalizes well to the external data set.

\section{Discussion}
We demonstrate that a simple \emph{DenseNet121} is able to serve as a surrogate model for the abstract human segmentation quality rating under a scarce training data regime.
Notably, the mean absolute error deviation is lower than the difference Kofler et al. \cite{kofler2021using} reported for individual human raters from the mean human rating.
Apparently, the 3D segmentation quality is sufficiently encoded in the 2D center of mass slices.
Our experiments show that the choice of hyperparameters is not critical as all networks reach solid performance.

Expert radiologists are among the highest-paid doctors and are notoriously hard to come by.
Given that the inference of the approximated quality rating only takes split seconds, there are broad potential applications for \emph{deep quality estimation (DQE)} networks:

One obvious application for \emph{DQE} is quality monitoring during inference.
Even though fully-automatic glioma segmentation algorithms, on average, tend to outperform human annotators \cite{kofler2021using},
they sometimes fail spectacularly.
For successful clinical translation of such algorithms, detection and mitigation of failure cases is imperative.

Another potential use case for \emph{DQE} is dat aset curation.
Data set curation is an important aspect of model training, as broken ground truth labels can destroy model performance.
As we demonstrate in the evaluation experiments, \emph{DQE} can differentiate between trustworthy and broken ground truth cases in a fully-automatic fashion.
This property becomes especially valuable in a federated learning setting, where researchers have no access to the ground truth labels.
\emph{DQE} allows training models only on trustworthy exams by applying a simple thresholding operation on the estimated quality score.

\noindent\textbf{Limitations:}
It is unclear how well our approach generalizes to other (segmentation) tasks and quality metrics.
Taking into account that glioma segmentation is a complex multi-class segmentation problem, the scarce training data, the 3D to 2D translation, and the abstract nature of the human-generated quality judgments, we believe there is a slight reason for optimism.

The proposed model performs better for predicting high-quality segmentations.
This is perhaps not surprising given that humans agree more on the quality of such cases and that these are better represented in the training data, as visible in \Cref{fig:network,fig:blanda,fig:pointo}.
Nevertheless, in practice, simple thresholding of the predicted star ratings can sufficiently distinguish segmentation qualities.

\noindent\textbf{Outlook:}
As we demonstrated, \emph{DQE} can approximate non-differentiable quality metrics, such as the abstract human segmentation quality rating, with a differentiable CNN.
This promises the possibility of training new (segmentation) networks with surrogates of non-differentiable quality metrics by using \emph{DQE} within the loss function.
\footnote{A big advantage here (compared to, e.g., \emph{GAN} training) is the possibility to train the networks sequentially and thereby stabilize the training process.}
Future research should address these open questions.

\clearpage

\section*{Acknowledgement}
\noindent BM, BW and FK are supported through the SFB 824, subproject B12.

\noindent Supported by Deutsche Forschungsgemeinschaft (DFG) through TUM International Graduate School of Science and Engineering (IGSSE), GSC 81.

\noindent LF, SS, EDLR and IE are supported by the Translational Brain Imaging Training Network (TRABIT) under the European Union's `Horizon 2020' research \& innovation program (Grant agreement ID: 765148).

\noindent IE and SS are funded by DComEX (Grant agreement ID: 956201).

\noindent Supported by Anna Valentina Lioba Eleonora Claire Javid Mamasani.

\noindent With the support of the Technical University of Munich – Institute for Advanced Study, funded by the German Excellence Initiative.

\noindent EDLR is employed by ico\textbf{metrix} (Leuven, Belgium).

\noindent JP and SS are supported by the Graduate School of Bioengineering,  Technical University of Munich.

\noindent JK has received Grants from the ERC, DFG, BMBF and is Co-Founder of Bonescreen GmbH.

\noindent BM acknowledges support by the Helmut Horten Foundation.

\noindent Research reported in this publication was partly supported by the National Institutes of Health (NIH) under award numbers NIH/NCI:U01CA242871 and NIH/NINDS:R01NS042645.

\noindent Research reported in this publication was partly supported by AIME GPU cloud services.

\bibliographystyle{splncs04}
\bibliography{references}
% \addbibresource{references.bib}
%

\end{document}